\documentclass{article}
\usepackage{arxiv}
\usepackage[utf8]{inputenc} % allow utf-8 input
\usepackage[T1]{fontenc}    % use 8-bit T1 fonts
\usepackage{hyperref}       % hyperlinks
\usepackage{url}            % simple URL typesetting
\usepackage{booktabs}       % professional-quality tables
\usepackage{amsfonts}       % blackboard math symbols
\usepackage{nicefrac}       % compact symbols for 1/2, etc.
\usepackage{microtype}      % microtypography
\usepackage{graphicx} 
\usepackage{amsmath}
\usepackage{amssymb}
\usepackage{subcaption}
\usepackage{mwe}
\usepackage[sort, numbers]{natbib}
\usepackage{xcolor}
\title{SDM-Net: A Simple and Effective Model for Generalized Zero-Shot Learning}

\author{%
  Shabnam Daghaghi\\
  Electrical and Computer Engineering\\
  Rice University\\
  Houston, TX 77005 \\
  \texttt{shabnam.daghaghi@rice.edu} \\
   \And
   Tharun Medini \\
  Electrical and Computer Engineering \\
  Rice University\\
  Houston, TX 77005 \\
   \texttt{tharun.medini@rice.edu} \\
   \And
   Anshumali Shrivastava \\
  Department of Computer Science \\
  Rice University\\
  Houston, TX 77005 \\
   \texttt{anshumali@rice.edu} \\
}

\begin{document}

\maketitle

\begin{abstract}
  Zero-Shot Learning (ZSL) is a classification task where some classes referred to as \emph{unseen classes} have no training images. Instead, we only have side information about seen and unseen classes, often in the form of semantic or descriptive attributes. Lack of training images from a set of classes restricts the use of standard classification techniques and losses, including the widespread cross-entropy loss. We introduce a novel Similarity Distribution Matching Network (SDM-Net) which is a standard fully connected neural-network architecture with non-trainable penultimate layer consisting of class attributes. The output layer of SDM-Net consists of both seen and unseen classes. To enable zero-shot learning, during training, we regularize the model such that the predicted distribution of unseen class is close in KL divergence to the distribution of similarities between the correct seen class and all the unseen classes. We evaluate the proposed model on five benchmark datasets for zero-shot learning, AwA1, AwA2, aPY, SUN and CUB datasets. We show that, despite the simplicity, our approach achieves a competitive performance with state-of-the-art methods in Generalized-ZSL setting for all of these datasets

\end{abstract}

\section{Introduction}
Supervised classifiers, specifically Deep Neural Networks, need a large number of labeled samples to perform well. Deep learning frameworks are known to have limitations in fine-grained classification regime and detecting object categories with no labeled data ~\cite{xiao2015application,CrossmodelAndrewNG,goodbadugly,zhang2018zero}.
On the contrary, humans can recognize new classes using their previous knowledge. This power is due to the ability of humans to transfer their prior knowledge to recognize new objects ~\cite{fu2016semi,lake2015human}. Zero-shot learning aims to achieve this human-like capability for learning algorithms, which naturally reduces the burden of labeling. 

In zero-shot learning problem, there are no training samples available for a set of classes, referred to as unseen classes. Instead, semantic information (in the form of visual attributes or textual features) is available for unseen classes ~\cite{lampert2009learning,lampert2014attribute}. Besides, we have standard supervised training data along with the semantic information for a different set of classes, referred to as seen classes. The key to solve zero-shot learning problem is to train a classifier on seen classes to predict unseen classes by transferring knowledge analogous to humans. 

Early variants of ZSL assume that during inference, samples are only from unseen classes. Recent observations ~\cite{chao2016empirical,scheirer2013toward,goodbadugly} realize that such an assumption is not realistic. Generalized ZSL (GZSL) addresses this concern and considers a more practical variant. In GZSL there is no restriction on seen and unseen classes during inference. We are required to discriminate between all the classes. Clearly, GZSL is more challenging because the trained classifier is generally biased toward seen classes.

Where samples are image, in order to create a bridge between visual space (training data) and semantic attribute space (semantic information), some methods utilize embedding techniques ~\cite{TMitch2009,ESZSL,CrossmodelAndrewNG,bucher2016improving,xu2017transductive,zhang2017learning,kodirov2015unsupervised,akata2016label,akata2015evaluation,simonyan2014very,DeViSE,xian2016latent,zhang2016zero,al2016recovering,pmlr-v97-zhang19l,DBLP:journals/corr/abs-1812-09903} and the others use semantic similarity between seen and unseen classes ~\cite{zhang2015zero,fu2015zero,mensink2014costa}. Semantic similarity based models represent each unseen class as a mixture of seen classes. While the embedding based models follow three various directions;
mapping visual space to semantic space ~\cite{TMitch2009,ESZSL,CrossmodelAndrewNG,bucher2016improving,xu2017transductive,CrossmodelAndrewNG}, mapping semantic space to the visual space ~\cite{zhang2017learning,kodirov2015unsupervised,shojaee2016semi,ye2017zero}, and mapping both visual and semantic space into a joint embedding space~\cite{akata2016label,akata2015evaluation,simonyan2014very,DeViSE,xian2016latent,zhang2016zero,al2016recovering}. 
 
% The loss functions in embedding based models have training samples only from the seen classes. For unseen classes, we do not have any samples. It is not difficult to see that this lack of training samples biases the learning process towards seen classes only. One of the recently proposed techniques to address this issue is augmenting the loss function with some unsupervised regularization such as entropy minimization over the unseen classes ~\cite{liu2018generalized}.

Another recent methodology which follows a different perspective is deploying Generative Adversarial Network (GAN) to generate synthetic samples for unseen classes by utilizing their attribute information ~\cite{mishra2018generative,zhu2018generative,xian2018feature,felix2018multi,kumar2018generalized}. Although generative models boost the results significantly, it was argued that they are more difficult to train ~\cite{sutskever2015towards,salimans2016improved}. Furthermore, training requires generation of large number of samples followed by training on a much larger augmented data which hurts their scalability.

A recent notable model called DCN ~\cite{liu2018generalized}, is also based on mapping visual features and semantic attributes to a common embedding. DCN minimizes cross-entropy on seen classes to learn their visual features. For the model to not ignore the unseen class attributes, it employs a regularization. DCN chooses to minimize the entropy of unseen classes. In essence, it is forcing the network not to predict uniform distribution (maximum entropy). Instead, it forces all the probability mass on one of the unseen class (least entropy). While this entropy regularization is simple and remarkably improves network accuracy on unseen classes, we argue that it is sub-optimal. Consider an example where the correct class is a squirrel, and we have rats, mice, and several similar rodents in the unseen class. These unseen classes likely have attributes similar to the squirrel's attribute. However, forcing the network to concentrate the probability only on one class is likely losing information. Worse, nothing stops the entropy loss to converge on an utterly wrong class, which also minimized the entropy. 

To resolve these issues, we propose a Similarity Distribution Matching (SDM) regularizer, which enforces a complete distribution on the unseen class to match the distribution obtained from the semantic similarity of class attributes with the correct class. SDM therefore uses attribute information in a more explicit way. The regularization imposes a larger structure on the network. We show that SDM outperforms DCN by a significant margin. %and achieves state-of-the-art performance on ZSL benchamrk datasets.

\subsection{Our Contribution}
We propose a simple, fully connected neural network architecture with unified (both seen and unseen classes together) cross-entropy loss. Our proposal differs from a standard neural network for supervised classification in two ways. The first difference is that in the proposed network, the final layer is fixed and non-trainable. The weight vectors for neurons in the last non-trainable layer are precisely the available semantic attributes. We argue that this standard architecture is no less powerful than the popular embedding models.  The second difference is a novel loss function based on semantic similarity-based regularization.

In ZSL, due to lack of training data for the unseen class, after minimizing any loss function over the training data, the classifier will always prefer seen classes over unseen classes. This is the main challenge of ZSL problem where for any given input, the predicted class will likely only come from the seen classes. We propose Similarity Distribution Matching (SDM) to regularize this minimization problem which enables training data from seen classes to also learn and even predict unseen classes.

In particular, we directly use attribute similarity information between the correct seen class and the unseen classes to create a regularizer. Among all classifiers with small training loss on the seen data, we prefer classifiers whose predicted probability distribution on unseen classes, matches the normalized similarity distribution. The similarity distribution is defined by the attribute similarity between the correct seen class and all the unseen classes.  As a result of SDM, training instances for seen classes also serve as proxy training instances for the unseen class without increasing the training corpus. This SDM after simplification leads to a straightforward regularizer, which we argue is more informative than the recently proposed entropy regularizer ~\cite{liu2018generalized}. SDM regularization, along with cross-entropy loss, enables a simple MLP network to tackle GZSL problem. Our proposed model achieves a competitive performance with state-of-the-art methods in Generalized-ZSL setting on all five ZSL benchmark datasets.

\section{Related Works}
The main goal in ZSL problem is to bridge the gap between visual features and semantic representations of unseen classes. Semantic representations are usually available in the form of word embeddings learned on text corpus or human annotation attributes. Some early ZSL methods utilized a two-step approach, ~\cite{lampert2009learning,al2016recovering} learn a probabilistic attribute classifier and then estimate class posteriors, ~\cite{norouzi2013zero} predict seen class posteriors then take the convex combination of class label embeddings to project images into the semantic space. These two-step methods suffer from projection domain shift ~\cite{Fu:2015:TMZ:2881666.2882202}. On the other hand, recent works in ZSL directly learn an embedding between visual and semantic representations. ~\cite{akata2015evaluation,akata2013label} learn a bilinear compatibility function through structural SVM loss and ranking loss, respectively.
DeViSE \cite{DeViSE} utilizes a pairwise ranking loss to learn a mapping between visual space and semantic space. ESZSL \cite{ESZSL} introduces a simple analytical approach and utilizes square loss with $L_2$ norm regularization to learn the compatibility function between visual and semantic space. 

Loss functions in embedding based models have training samples only from seen classes and there is no sample from unseen classes. It is not difficult to see that this lack of training samples biases the learning process towards seen classes only. One of the recently proposed techniques to address this issue is augmenting the loss function with some unsupervised regularization such as entropy minimization over the unseen classes ~\cite{liu2018generalized}.

The two most recent state-of-the-art discriminative GZSL methods, CRnet ~\cite{pmlr-v97-zhang19l} and COSMO ~\cite{DBLP:journals/corr/abs-1812-09903}, both employ a complex mixture of experts approach. CRnet is based on k-means clustering with an expert module on each cluster (seen class) to map semantic space to visual space. The output of experts (cooperation modules) are integrated and finally sent to a complex loss (relation module) to make a decision. CRnet is a multi-module (multi-network) method that needs end-to-end training with many hyperparameters. Also COSMO is a complex gating model with three modules: a seen/unseen classifier and two expert classifiers over seen and unseen classes. Both of these methods have many modules, and hence, several hyperparameters; architectural, and learning decisions. A complex pipeline is susceptible to errors, for example, CRnet uses k-means clustering for training and determining the number of experts and a weak clustering will lead to bad results.

Utilizing generative models is a totally different approach in GZSL setting. Given semantic representation of classes, GANs aim to synthesize visual features and turn GZSl problem into a conventional classification problem. ~\cite{xian2018feature} generates discriminative visual features through paring Wasserstein GAN ~\cite{ishaan2017improved} with classification loss. ~\cite{Sariyildiz_2019_CVPR} extends this notion with gradient matching loss and learning an unconditional discriminator. ~\cite{mishra2018generative} train a conditional Variational Auto-Encoder (cVAE) to generates samples from given semantic representations. ~\cite{kumar2018generalized} follows similar approach and adds a multivariate regressor to map the generated samples to the relevant semantic representation. ~\cite{xian2019f} employs both GAN and VAE with an additional discriminator to generate more discriminative features.   

Our proposed model follows a discriminative framework to solve GZSL setting. We map visual features to semantic space, calculate similarity measure in semantic space and finally apply a Softmax classifier. By utilizing SDM regularization, We efficiently implement all three components by a simple MLP network.

\section{Problem Definition}
% In zero-shot learning problem, a set of training data on seen classes and a set of semantic information (attributes) on both seen and unseen classes are given. 
Training dataset $\mathcal{D}=\left \{ \left ( \textbf{x}_i,\textbf{y}_i \right ) \right \}_{i=1}^{n}$ includes $n$ samples where $\textbf{x}_i$ is the visual feature vector of the $i$-th image and $\textbf{y}_i$ is the class label. All samples in $\mathcal{D}$  belong to seen classes $\mathcal{S}$ and during training there is no sample available from unseen classes $\mathcal{U}$. The total number of classes is $C=|\mathcal{S}|+|\mathcal{U}|$. Semantic information or attributes $\textbf{a}_k \in \mathbb{R}^{a}$, are given for all $C$ classes and the collection of all attributes are represented by attribute matrix $\textbf{A}\in \mathbb{R}^{a\times C}$.
In the inference phase, our objective is to predict correct classes for the test dataset $\mathcal{D}'$. Classic ZSL setting assumes that all test samples in $\mathcal{D}'$ belong to unseen classes $\mathcal{U}$ and tries to classify test samples only to unseen classes $\mathcal{U}$. While in a more realistic setting i.e. GZSL, there is no such an assumption %about correct classes of test data $\mathcal{D}'$ %
and we aim at classifying samples in $\mathcal{D}'$ to either seen or unseen classes $\mathcal{S}\cup \mathcal{U}$.

In the next few sections, we outline the specific components of our method. Proposed network architecture, SDM regularization details, and the employed training strategy are presented.
\subsection{Network Architecture}

As Figure \ref{NNdiagram} illustrates our architecture, the architecture is a fully connected neural network that takes visual features and predicts the class using standard Softmax and Similarity Distribution Matching (described later). The final Softmax layer has one node for every class, both seen and unseen. To incorporate the semantic attributes of every class in the learning process, we replace the weight corresponding to each class node, in the Softmax layer, with the given attributes and make this layer non-trainable or immutable during training. The penultimate layer of the network has the same size as the dimension of semantic embedding of the class. 

\subsubsection{Why is this Architecture Sufficient?}

We contrast our approach with popular embedding based approaches for zero-shot learning and argue that the two approaches are equally powerful in terms of architecture. 

The whole learning process is about finding the right compatibility score between the input visual features, call it $v$, and the class attributes, call it $a$.  In embedding models, this score is computed via inner product between embedding of visual features and the embedding of attributes. Lets $f(v)$ denotes the embedding of $v$ and $g(a)$ denotes embedding of $a$. Here, $f$ and $g$ are non-linear functions which are generally a neural networks. Thus, the compatibility between $v$ and $a$, call it $C(v,a)$ is given by
$$C(v,a) = \langle f(v) , g(a)\rangle$$
In our architecture, the score is Softmax which is monotonic in the inner product of the penultimate embedding, which is a function of $v$ (call it $f'(v)$), and the semantic attributed $a$. Thus, with our model we can write 
$$C(v,a) = \langle f'(v) , a\rangle$$
Since $f'$ is non-linear neural network and can be as complex as we want, given $f$ and $g$, we can always choose $f'$ complex enough such that 
$$\langle f'(v) , a\rangle = \langle f(v) , g(a)\rangle$$

As a result, the power of our architecture is no less than the power of standard embedding models. Instead, we believe that  Softmax is more powerful modelling. Unlike embedding models which are limited to modeling pairwise compatibility, Softmax models the complete conditional distribution.  We will be precisely needing the information of the complete joint distribution to propose a superior regularizer. Besides, the simple Softmax and its standard probability interpretation will eliminate the need for fancy normalization which otherwise is a concern for embedding models. 

\subsection{Similarity Distribution Matching}
In ZSL problem,
%there are no training instances from unseen classes, so 
the network output nodes corresponding to unseen classes are always inactive during learning
%Standard supervised training with cross-entropy loss biases the network towards seen classes only. The true labels used for training only represent seen classes and 
since cross-entropy loss cannot penalize unseen classes. Moreover, the available similarity information between seen and unseen attributes is never utilized explicitly.

We overcome this inherent bias by regularizing the network to reproduce a predetermined probability distribution on unseen classes where this probability is dictated by semantic similarity. We propose creating unseen probability distribution based on the similarity between semantic attributes. For each seen sample, we represent its relationship to unseen categories by obtaining the semantic similarity (dot-product) of its attribute with the attributes of unseen classes.

We then squash all these dot-product values by Softmax to acquire probabilities (Equation \ref{eq:softl2}). While training, we use this similarity distribution to regularize the classifier. In particular, we enforce the predicted probability distribution on the unseen class close to the prescribed similarity distribution. 

In order to control the flatness of the unseen distribution, we utilize temperature parameter $\tau$ in Softmax ~\cite{hinton2015distilling}.  
%The temperature parameter $\tau$, controls the flatness of the unseen distribution;
Higher temperature results in flatter distribution over unseen classes and lower temperature creates a more ragged distribution with peaks on nearest unseen classes.
%A small enough temperature basically results in the nearest unseen approach.
The impact of temperature $\tau$ on unseen distribution is depicted in Figure \ref{fig:temp_effect}.\textit{left} for a particular seen class. SDM regularizer implicitly introduces unseen visual features into the network without generating fake unseen samples as in generative methods ~\cite{mishra2018generative,zhu2018generative,xian2018feature}.
%Hence our proposed approach is able to reproduce same effect as in generative models without the need to create fake samples and train generative models which are known to be difficult to train.
Below is the formal description of  \textit{temperature} Softmax to produce similarity distribution of unseen class $k$ for seen class $i$ (unseen probability distribution):
\begin{equation}
\label{eq:softl2}
y_{i,k}^u = q \frac{\text{exp} \left (s_{i,k}/\tau  \right )}{ \sum_{j \in \mathcal{U}} \text{exp} \left ( s_{i,j}/\tau  \right )} \; \; \; \; \; \text{where} \; \; \; \;  
s_{i,j}\triangleq  \langle \textbf{a}_i ,\textbf{a}_j \rangle
\end{equation}
where $\textbf{a}_i$ is the $i$-th column of attribute matrix $\textbf{A}\in \mathbb{R}^{a\times C}$ which includes both seen and unseen class attributes: $\textbf{A} = \begin{bmatrix} \textbf{a}_1 \;|\; \textbf{a}_2 \;|\; \cdots \;|\; \textbf{a}_C
\end{bmatrix}$. And $s_{i,j}$ is the \textit{true} similarity score between two classes $i,j$ based on their attributes. $\tau$ and $q$ are temperature parameter and unseen similarity distribution regularization factor, respectively.

\begin{figure}
\centering
\includegraphics[scale=0.6]{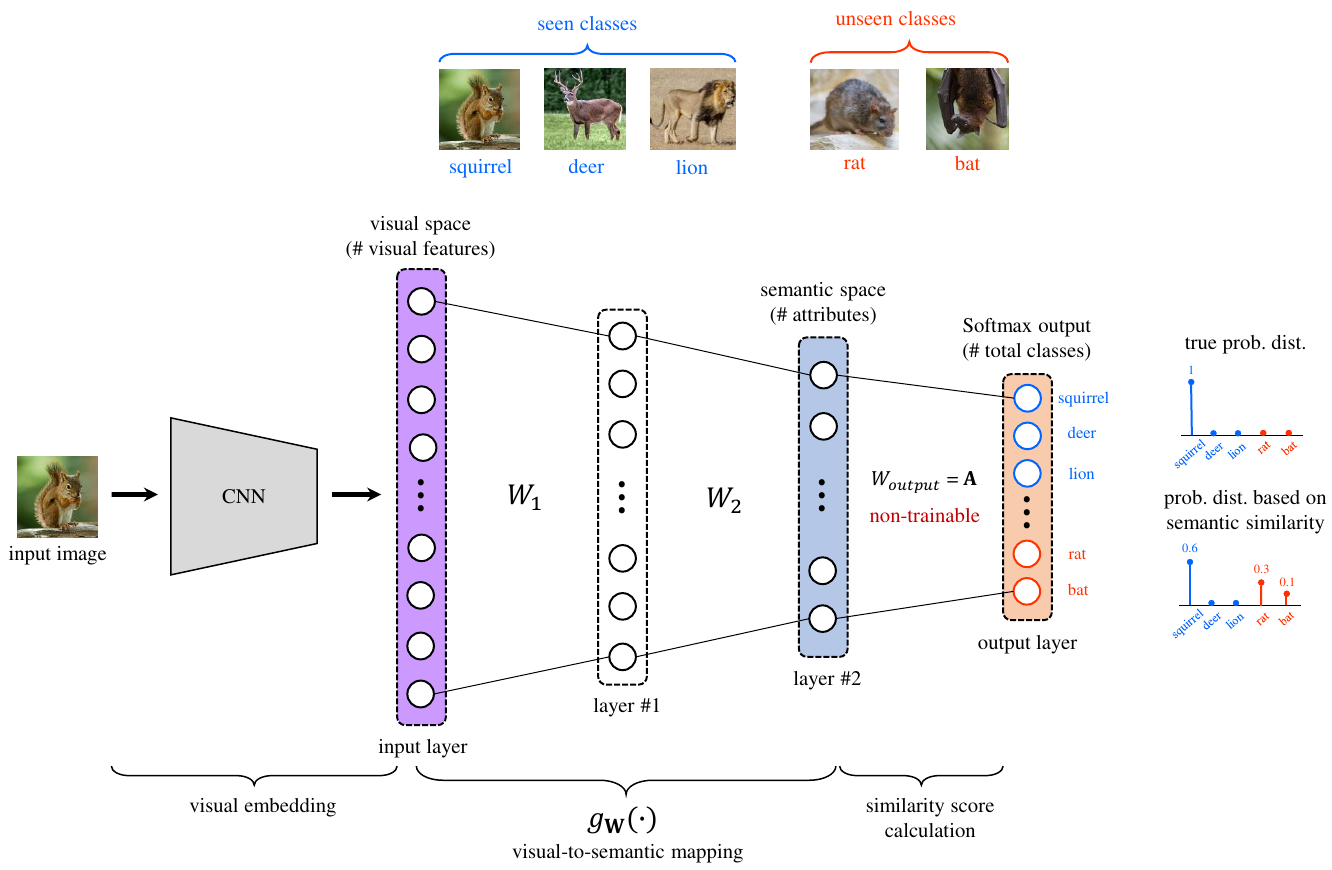}
% \vspace{3mm}
\caption{Overall workflow of the SDM classifier and architecture of the proposed MLP. Layers \#1 and \#2 provide the nonlinear embedding $g_\textbf{W}(.)$ to map visual features to attribute space and their weights $W_1$, $W_2$ are learned by SGD. The output layer with non-trainable weights $\textbf{A}$, basically calculates dot-products of semantic representation of the input and all class attributes simultaneously. Probability distribution based on semantic similarity is also shown for a sample image from \textit{squirrel} class.}
\label{NNdiagram}
\end{figure}

% \begin{figure}
% \centering
% \includegraphics[scale=0.7]{overal_diagram_05212019.pdf}
% \caption{The overall framework of the proposed Z-Softmax classifier. Semantic representation $\Tilde{a}$ is obtained via a visual-to-semantic mapping of visual features $\textbf{x}$. Similarity scores (dot-products) of $\Tilde{a}$ and all class attributes $\textbf{A}$ are calculated and passed through a Softmax to produce all class probabilities.}
% \label{overaldiagram}
% \end{figure}

% \begin{figure}
% \centering
% \includegraphics[scale=0.8]{NN_diagram_05212019.pdf}
% \caption{Architecture of the proposed MLP for Z-Softmax classifier. Layers \#1 and \#2 provide the nonlinear embedding $g_\textbf{W}(.)$ to map visual features to attribute space and their weights $W_1$, $W_2$ are learned by SGD. The output layer with non-trainable weights $\textbf{A}$, basically calculates dot-products of semantic representation of the input and all class attributes simultaneously. }
% \label{NNdiagram}
% \end{figure}

The proposed method is a multi-class probabilistic classifier that produces a $C$-dimensional vector of class probabilities $\textbf{p}$ for each sample $\textbf{x}_i$ as $\textbf{p}(\textbf{x}_i)=\textit{Softmax} \left ( \textbf{A}^Tg_\textbf{w}\left ( \textbf{x}_i \right ) \right )$ where $\textbf{A}^Tg_\textbf{w}\left ( \textbf{x}_i \right )$ is a $C$-dimensional vector of all similarity scores for an input sample (Figure \ref{NNdiagram}).

A natural choice to train such classifiers is the cross-entropy loss which we later show naturally integrates our idea of similarity distribution matching. 
Therefore, the optimization problem of our framework is as:
% During training, we aim at learning the nonlinear mapping $g_\textbf{w}(.)$ i.e. obtaining network weights $\textbf{W}$ through:
\begin{equation}
\label{eq:OF}
\underset{\textbf{W}}{\text{min}} \; \sum_{i=1}^{n}\mathcal{L}(\textbf{x}_i) + \lambda \left \| \textbf{W} \right \|_F^2
\end{equation}
where $\lambda$ is the weight decay regularization factor, and $\mathcal{L}(\textbf{x}_i)$
is the cross-entropy loss over true probability distribution ($L$) regularized by cross-entropy over probability distribution based on semantic similarity ($R$) for each sample as shown below:
 %%%%
\begin{equation}
\label{eq:Li1}
% L(\textbf{x}_i)=\sum_{k=1}^{C} y_{i,k}\text{log}(p_k(\textbf{x}_i))
% \centering
\mathcal{L}(\textbf{x}_i) = (1-\alpha) L(\textbf{x}_i)  + \alpha R(\textbf{x}_i)%=\alpha \sum_{k\in C} y_{k}^{soft}\text{log}(p_k) + (1-\alpha) \sum_{k\in C} y_{k}^{hard}\text{log}(p_k) 
\end{equation}

where $\alpha \in [0,1]$ is SDM regularization parameter. Through $R$, SDM regularizer, we want to regularize the overconfidence of classifier toward seen classes and enrich the network with the ability to also identify unseen samples. 

The regularizer term can be expanded to seen and unseen terms as follows (omitting $i$ subscript for simplicity):
\begin{equation}
\label{eq:Li2}
% \sum_{k\in C} y_{k}^{soft}\text{log}(p_k)= \sum_{k\in C_s} y_{k}^s\text{log}(p_k^s)+\sum_{k\in C_u} y_{k}^u\text{log}(p_k^u)  
R(\textbf{x}) = -\sum_{k\in \mathcal{S}} y_{k}^s\text{log}(p_k^s)-\sum_{k\in \mathcal{U}} y_{k}^u\text{log}(p_k^u) 
\end{equation}
where $y_k^s$ and $y_k^u$ are probability distributions (based on semantic similarity) for seen and unseen class $k$, respectively. $y_k^u$ is in fact the unseen similarity distribution (Equation \ref{eq:softl2}) that the model attempts to match via SDM regularizer. The second term of Equation \ref{eq:Li2} is KL divergence (omitting the constant entropy term on $y_k^u$) that matches similarity distribution of unseen classes and their corresponding predicted probability distribution.

We observe that the SDM  regularizer is a weighted cross-entropy on unseen class, which leverages similarity structure between attributes as opposed to uniform entropy function of DCN ~\cite{liu2018generalized}. DCN and all prior works use uniform entropy as regularizer, which does not capitalize on the known semantic similarity information between seen and unseen class attributes.

At the inference time, our proposed SDM method works the same as a conventional classifier, we only need to provide the test image and the network will produce class probabilities for all seen and unseen classes.

\section{Experiment}
We conduct comprehensive comparison of our proposed SDM model with the state-of-the-art discriminative methods for GZSL settings on five benchmark datasets (Table \ref{table1}). Our model achieves a competitive performance with the state-of-the-art methods on GZSL setting for all benchmark datasets. We present the details in the following.

\begin{table}[t]
% \vspace{-0.3in}
% \tiny
% \scriptsize

%   \centering
\begin{center}
  \begin{tabular}{lllll}
    % \toprule
    %\multicolumn{4}{c}{AwA}                   \\
    % \cmidrule(r){1-4}
    \hline
    Dataset   &\# Attributes    &\# Seen Classes      &\# Unseen Classes  &\# Images \\
    % \midrule
    \hline
    AwA1     &85 &40   &10 &30475 \\
     AwA2     &85 &40   &10 &37322  \\
    CUB     &312  &150    &50 &11788 \\
  aPY     &64  &20   &12 &18627 \\
    SUN   &102  &645  &72  &14340 \\
   \hline
    \bottomrule
  \end{tabular}
  \end{center}
  \caption{Statistics of five ZSL benchmark datasets}
  \label{table1}
\end{table}

\subsection{Dataset}
The proposed method is evaluated on five benchmark ZSL datasets. The statistics for the datasets are shown in Table \ref{table1}. Animal with Attributes (AwA1) ~\cite{lampert2014attribute} dataset is a coarse-grained benchmark dataset for ZSL/GSZl. It has 30475 image samples from 50 classes of different animals and each class comes with side information in the form of attributes (e.g. animal size, color, place of habitat). Attribute space dimension is 85 and this dataset has a standard split of 40 seen and 10 unseen classes introduced in ~\cite{lampert2014attribute}. AWA2 ~\cite{goodbadugly} is the public licensed version of AWA1 with roughly the same amount of samples and the same number of attributes and seen/unseen classes as AWA1. 

Caltech-UCSD-Birds-200-2011 (CUB) ~\cite{WahCUB_200_2011} is a fine-grained ZSL benchmark dataset. It has 11,788 images from 200 different types of birds and each class comes with 312 attributes. The standard ZSL split for this dataset has 150 seen and 50 unseen classes ~\cite{akata2016label}.

SUN Attribute (SUN) ~\cite{patterson2012sun} is a fine-grained ZSL benchmark dataset consists of 14340 images of different scenes and each scene class is annotated with 102 attributes. This dataset has a standard ZSL split of 645 seen and 72 unseen classes.

attribute Pascal and Yahoo (aPY) ~\cite{farhadi2009describing} is a small and coarse-grained ZSL benchmark dataset which has 14340 images and 32 classes of different objects (e.g. aeroplane, bottle, person, sofa, ...) and each class is provided with 64 attributes. This dataset has a standard split of 20 seen classes and 12 unseen classes. 

\subsection{Evaluation Metric}

For the purpose of validation, we employ the validation splits provided along with the Proposed Split (PS) ~\cite{goodbadugly} to perform cross-validation for hyper-parameter tuning. The main objective of GZSL is to simultaneously improve seen samples accuracy and unseen samples accuracy i.e. imposing a trade-off between these two metrics. As the result, the standard GZSL evaluation metric is harmonic average of seen and unseen accuracy. This metric is chosen to encourage the network not to be biased toward seen classes. Harmonic average of accuracies is defined as $A_H = \frac{2A_SA_U}{A_S+A_U}$ where $A_S$ and $A_U$ are seen and unseen accuracies, respectively.

\begin{table*}[h]
%  \tiny
\scriptsize
% \footnotesize
% \captionsetup[table]{skip=10pt}

%   \centering
\begin{center}
  \begin{tabular}{p{3.5cm}|p{0.22cm}p{0.22cm}p{0.27cm}|p{0.22cm}p{0.22cm}p{0.27cm}|p{0.22cm}p{0.22cm}p{0.27cm}|p{0.22cm}p{0.22cm}p{0.27cm}|p{0.22cm}p{0.22cm}p{0.27cm}}
    \toprule
   \hline
    \multicolumn{1}{c}{}
      &\multicolumn{3}{c}{AwA1}
    &\multicolumn{3}{c}{AwA2}
    &\multicolumn{3}{c}{aPY}
    &\multicolumn{3}{c}{CUB}
    &\multicolumn{3}{c}{SUN}
     
    \\
        % \midrule
    % \cmidrule(r){1-4}
    
    Method   &U   &S      &H &U  &S  &H &U  &S  &H &U  &S  &H &U  &S  &H \\
    % \midrule
    \hline
    % \textbf{Non-Generative Models} &{} &{} &{} &{} &{} &{} &{} &{} &{} &{} &{} &{}\\
    DAP ~\cite{lampert2009learning}  & 0.0  &\textbf{88.7}   
    &0.0 &- &- &- & 4.8 &78.3 &9.0 &1.7 &67.9 &3.3  &4.2 &25.1 &7.2\\
    ALE ~\cite{akata2013label} & 16.8  & 76.1   
    &27.5 &14.0 &81.8  &23.9 &4.6 &73.7 &8.7 &23.7 &62.8 &34.4  &21.8 &33.1 &26.3\\
    SJE ~\cite{akata2015evaluation}     &11.3 & 74.6   &19.6  &8.0 &73.9  &14.4  &3.7 &55.7 &6.9 &23.5 &59.2 &33.6  &14.7 &30.5 &19.8      \\
    % LATEM ~\cite{xian2016latent}     &7.3 & 71.7   &13.3  &11.5 &77.3 &20.0  &0.1  &73.0 &0.2 &15.2 &57.3 &24.0 &14.7 &28.8 &19.5     \\
    SSE ~\cite{zhang2015zero}  &7.0  &80.5 &12.9 &8.1 &82.5 &14.8 &0.2 &78.9  &0.4  &8.5 &46.9 &14.4 &2.1 &36.4 &4.0 \\ 
    ConSE ~\cite{norouzi2013zero}     &0.4  &88.6    &0.8  &0.5 &\textbf{90.6}  &1.0   &0.0 &\textbf{91.2} &0.0 &1.6 &\textbf{72.2} &3.1 &6.8 &39.9 &11.6  \\
    Sync ~\cite{changpinyo2016synthesized}     &8.9  &87.3   &16.2  &10.0 &90.5  &18.0  &7.4 &66.3 &13.3 &11.5 &70.9 &19.8 &7.9 &\textbf{43.3} &13.4\\
    ESZSL ~\cite{ESZSL}  &6.6 &75.6 &12.1 &5.9 &77.8 &11.0  &2.4 &70.1 &4.6 &12.6 &63.8 &21.0  &11.0 &27.9 &15.8\\ 
    DeViSE ~\cite{DeViSE}   &13.4  &68.7  &22.4 &17.1 &74.7  &27.8  &4.9 &76.9 &9.2 &23.8 &53.0 &32.8 &16.9 &27.4 &20.9\\
    CMT ~\cite{CrossmodelAndrewNG} &0.9  &87.6  &1.8  &8.7 &89.0  &15.9 &1.4 &85.2 &2.8 &7.2 &49.8 &12.6 &8.1 &21.8 &11.8\\
    % \midrule
    % \hline
    %  \textbf{Generative Models} &{} &{} &{} &{} &{} &{} &{} &{} &{} &{} &{} &{}\\
    %     f-CLSWGAN ~\cite{xian2018feature} &57.9  &61.4 &59.6 &- &- &- &- &- &- &43.7 &57.7 &49.7 &42.6 &36.6 &39.4  \\
    %     % RN \citep{sung2018learning}    &31.4 &91.3 &46.7 &30.0 &93.4  &45.3 &- &- &-  &38.1 &61.1  &47.0 &- &- &-\\
    %  SP-AEN ~\cite{chen2018zero}  &23.3  &90.9 &37.1 &- &- &- &13.7 &63.4 &13.7 &34.7 &70.6 &46.6   &24.9 &38.6 &30.3  \\
    % cycle-UWGAN ~\cite{felix2018multi}  &59.6  &63.4 &59.8 &- &- &- &- &- &- &47.9 &59.3 &53.0 &47.2 &33.8 &39.4  \\
    %  SE-GZSL ~\cite{kumar2018generalized}  &56.3 &67.8 &61.5 &- &- &- &- &- &- &46.7 &53.3 &41.5 &40.9 &30.5 &34.9  \\
    % %  CADA-VAE \citep{?}  &57.3 &72.8  &64.1  &55.8 &75.0 &63.9 &- &- &-  &51.6 &53.5 &52.4 &47.2 &35.7  &40.6\\
    % %  \midrule
    %  \hline
     ZSKL ~\cite{zhang2018zero}     &18.9 &82.7  &30.8 &- &- &-  &10.5 &76.2 &18.5 &21.6 &52.8 &30.6 &20.1 &31.4 &24.5 \\
           \hline
     DCN ~\cite{liu2018generalized}     &25.5  &84.2 &39.1 &- &- &-  &14.2 &75.0 &23.9 &28.4 &60.7 &38.7 &25.5 &37.0 &30.2 \\
     COSMO ~\cite{DBLP:journals/corr/abs-1812-09903}    &52.8  &80.0 &63.6   &- &- &-  &- &- &- &44.4 &57.8 &50.2 &44.9 &37.7 &\textbf{41.0}  \\
      CRnet ~\cite{pmlr-v97-zhang19l} &\textbf{58.1}  &74.7 &\textbf{65.4} &52.6	&52.6	&63.1  &32.4 &68.4 &44.0 &45.5 &56.8 &\textbf{50.5} &34.1 &36.5 &35.3   \\
        SDM-Net (Ours) &57.2   &75.8  &65.2 &\textbf{55.1}	&78.6	&\textbf{64.7}   &\textbf{42.7} &57.2 &\textbf{48.7} &\textbf{47.1} &52.5 &49.6 &\textbf{47.2} &32.6 &38.6  \\
   
    \hline
    \bottomrule

\end{tabular}
\end{center}
\vspace{3mm}
  \caption{Results of GZSL methods on ZSL benchmark datasets under Proposed Split (PS) \cite{goodbadugly}. U, S and H respectively stand for Unseen, Seen and Harmonic average accuracies.}
   \label{table22}
\end{table*}

\subsection{Implementation Details}

To evaluate SDM, we  follow the popular experimental framework and the Proposed Split (PS) in ~\cite{goodbadugly} for splitting classes into seen and unseen classes to fairly compare GZSL/ZSL methods. Utilizing PS ensures that none of the unseen classes have been used in the training of ResNet-101 on ImageNet. The input to the model is the visual features of each image sample extracted by a pre-trained ResNet-101 ~\cite{he2016deep} on ImageNet provided by ~\cite{goodbadugly}. The dimension of visual features is 2048. We do not fine-tune the CNN that generates visual features unlike model in \cite{liu2018generalized}. In this sense, our proposed model is also fast and straightforward to train.

We utilized Keras ~\cite{chollet2015} with TensorFlow back-end ~\cite{abadi2016tensorflow} to implement our model. We used ~\cite{goodbadugly} proposed unseen classes for validation (3-fold CV) and added 20\% of train samples (seen classes) as seen validation samples to obtain GZSL validation sets. We cross-validate $\tau\in[10^{-2},10]$, 
\emph{mini-batch size} $\in\{64,128,256,512,1024\}$, $q\in[0,1]$, $\alpha\in[0,1]$, $\lambda \in \{0,10^{-6},10^{-5},10^{-4}\}$,
\emph{hidden layer size} $\in\{128,256,512,1024,1500\}$ and \emph{activation function} $\in$\{tanh, sigmoid, hard-sigmoid, relu\} to tune our model. To obtain statistically consistent results, the reported accuracies are averaged over 5 trials (using different initialization) after tuning hyper-parameters with cross-validation. Also we ran our experiments on a machine with 56 vCPU cores, Intel(R) Xeon(R) CPU E5-2660 v4 @ 2.00GHZ and 2 NVIDIA-Tesla P100 GPUs each with 16GB memory. The code is provided in the supplementary material.

\begin{figure}
        \centering
        \vspace{-0.1in}
        \begin{subfigure}[b]{0.45\textwidth}
            \centering
            \includegraphics[width=\textwidth]{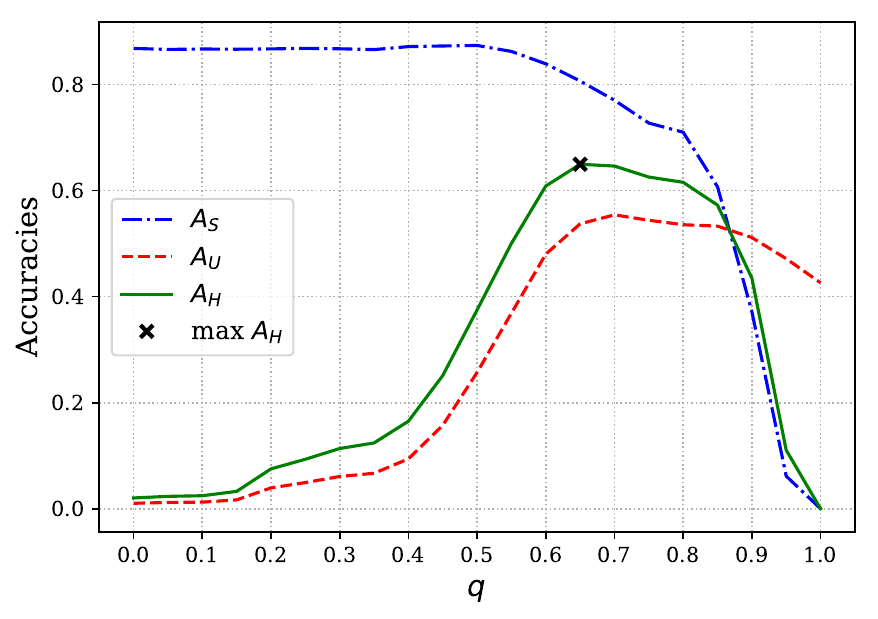}
            \caption[]%
            {{\small AwA}}    
            \label{}
        \end{subfigure}
        \begin{subfigure}[b]{0.45\textwidth}  
            \centering 
            \includegraphics[width=\textwidth]{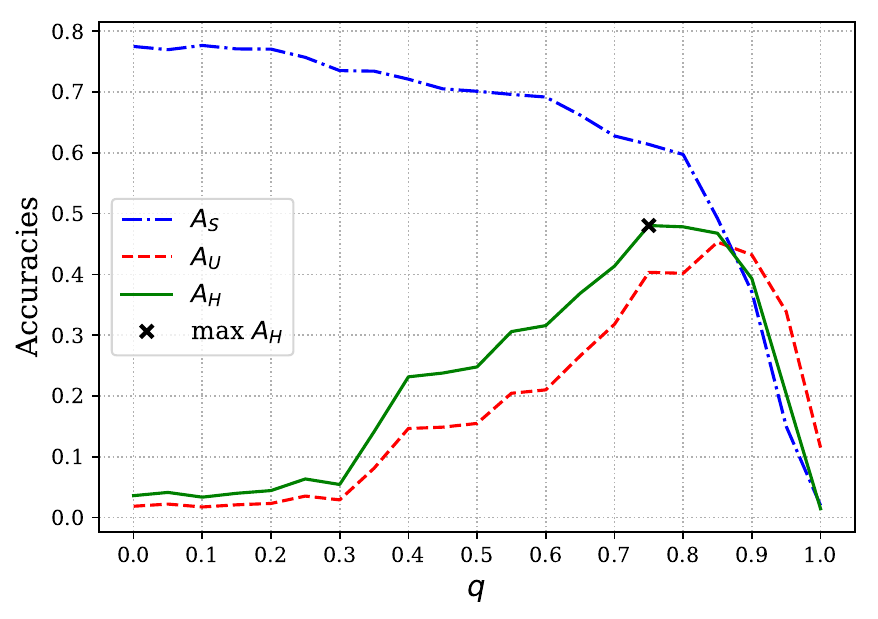}
            \caption[]%
            {{\small aPY}}    
            \label{}
        \end{subfigure}
        \begin{subfigure}[b]{0.45\textwidth}   
            \centering 
            \includegraphics[width=\textwidth]{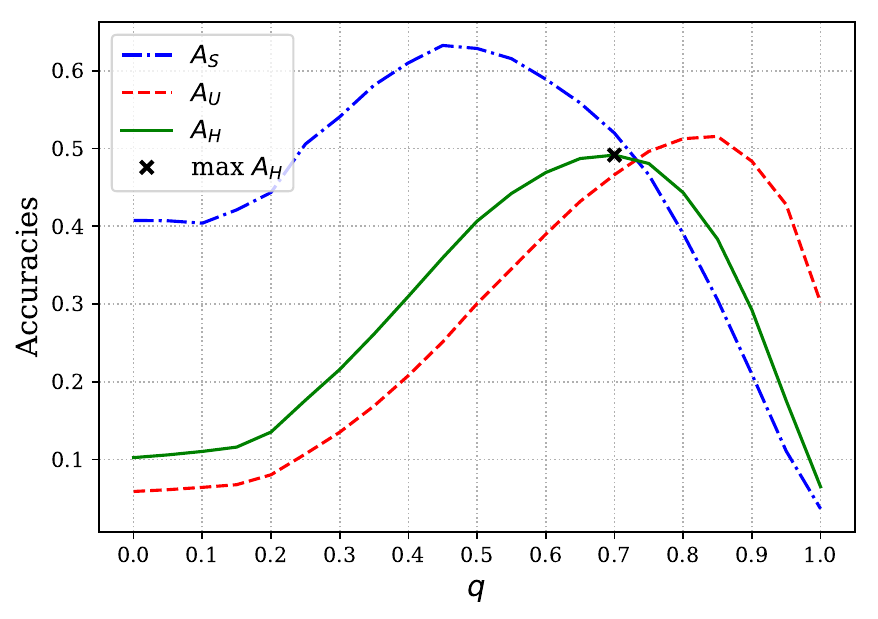}
            \caption[]%
            {{\small CUB}}    
            \label{}
        \end{subfigure}
        \begin{subfigure}[b]{0.45\textwidth}   
            \centering 
            \includegraphics[width=\textwidth]{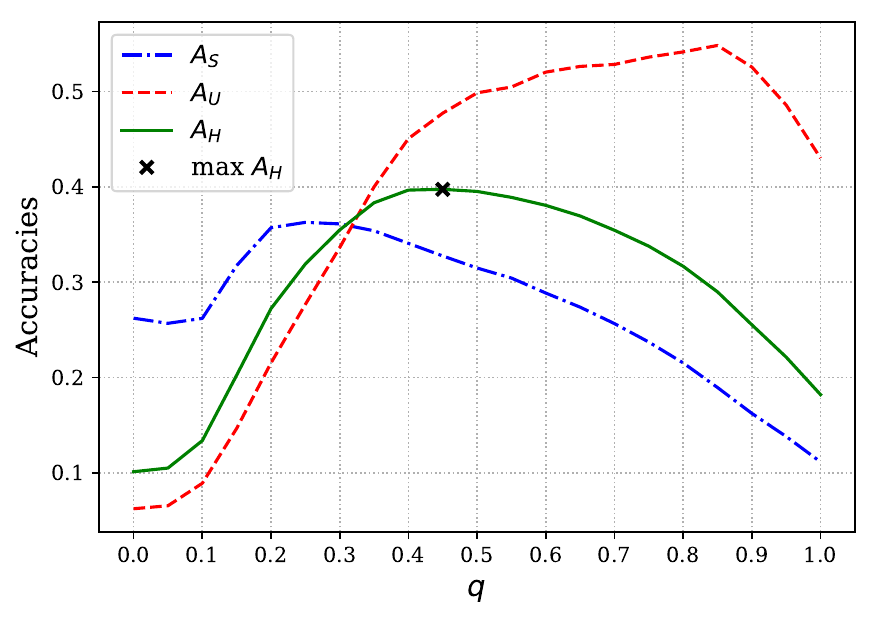}
            \caption[]%
            {{\small SUN}}    
            \label{}
        \end{subfigure}
        \caption[]
        {Plots of seen ($A_S$), unseen ($A_U$) and harmonic average ($A_H$) accuracies versus total probability ($q$) assigned to unseen classes are shown for all five datasets. The maximum obtained harmonic accuracy is also marked by ($\times $).} 
        \label{fig:acc_vs_q}
        \vspace{-0.2in}
    \end{figure}

\subsection{Generalized Zero-Shot Learning Results}
To demonstrate the effectiveness of SDM model in GZSL setting, we comprehensively compare our proposed method with state-of-the-art GZSL models in Table \ref{table22}. Since we use the standard proposed split, the published results of other GZSL models are directly comparable. 

As reported in Table \ref{table22}, accuracies of our model achieves the state-of-the-art GZSL performance on all five benchmark datasets and outperforms the state-of-the-art on AwA2 and aPY datasets. It is exciting and motivating while our architecture is much simpler compared to recently proposed  CRnet and COSMO, yet, we achieve similar or better accuracies compared to them. We have only one simple fully connected neural network with 2 trainable layers, compared to CRnet with K mixture of experts followed by relation module with complex loss functions (pairwise).

Semantic similarity distribution employed in SDM gives the model new flexibility to trade-off between seen and unseen accuracies during training and attain a higher value of harmonic accuracy $A_H$, which is the standard metric for GZSL. Assigned unseen probability ($q$) enables the classifier to gain more confidence in recognizing unseen classes, which in turn results in considerably higher unseen accuracy $A_U$. As the classifier is now discriminating between more classes we get marginally lower seen accuracy $A_S$. However, balancing $A_S$ and $A_U$ with the cost of deteriorating $A_S$ leads to much higher $A_H$. This trade-off phenomenon is depicted in Figure \ref{fig:acc_vs_q} for all datasets. The flexibility provided by SDM is examined by obtaining accuracies for different values of $q$. In Figure \ref{fig:acc_vs_q}.a and \ref{fig:acc_vs_q}.b, by increasing total unseen probability $q$, $A_U$ increases and $A_S$ decreases as expected. From the trade-off curves, there is an optimal $q$ where $A_H$ takes its maximum value as shown in Figure \ref{fig:acc_vs_q}. Maximizing $A_H$ is the primary objective in a GZSL problem that can be achieved by semantic similarity distribution matching and the trade-off knob, $q$. 

\begin{figure}
        \centering
        \vspace{-0.1in}
        \begin{subfigure}[b]{0.45\textwidth}
            \centering
            \includegraphics[width=\textwidth]{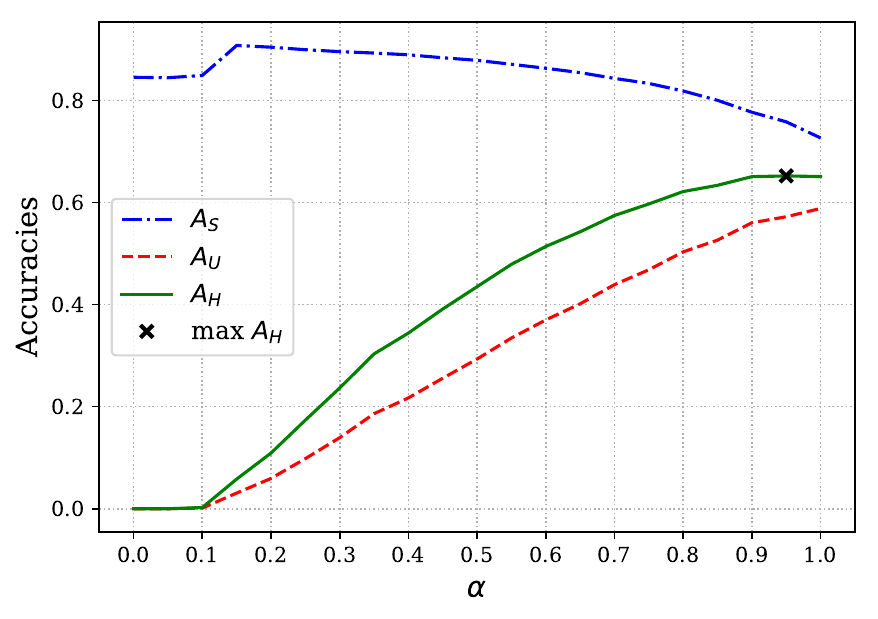}
            \caption[]%
            {{\small AwA}}    
            \label{}
        \end{subfigure}
        \begin{subfigure}[b]{0.45\textwidth}  
            \centering 
            \includegraphics[width=\textwidth]{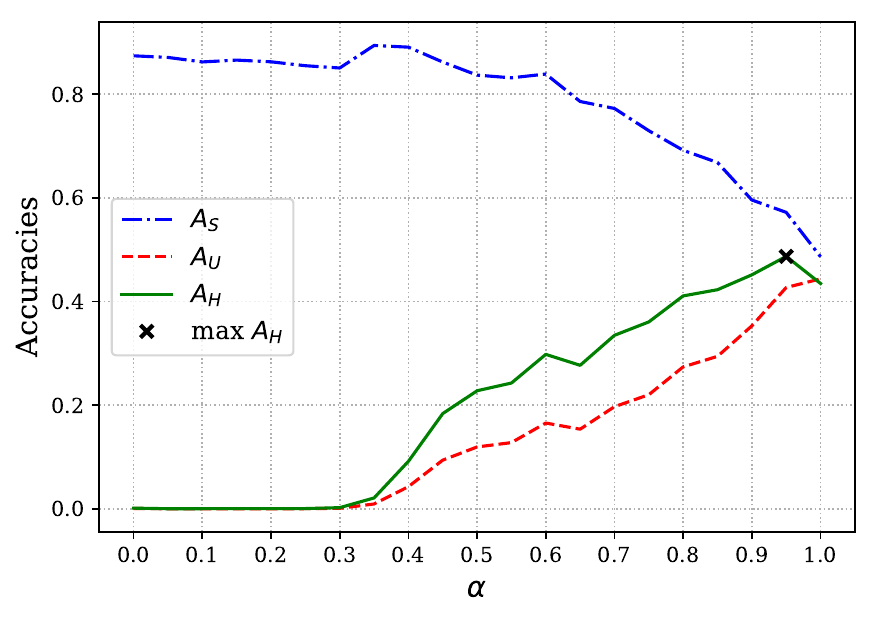}
            \caption[]%
            {{\small aPY}}    
            \label{}
        \end{subfigure}
        \begin{subfigure}[b]{0.45\textwidth}   
            \centering 
            \includegraphics[width=\textwidth]{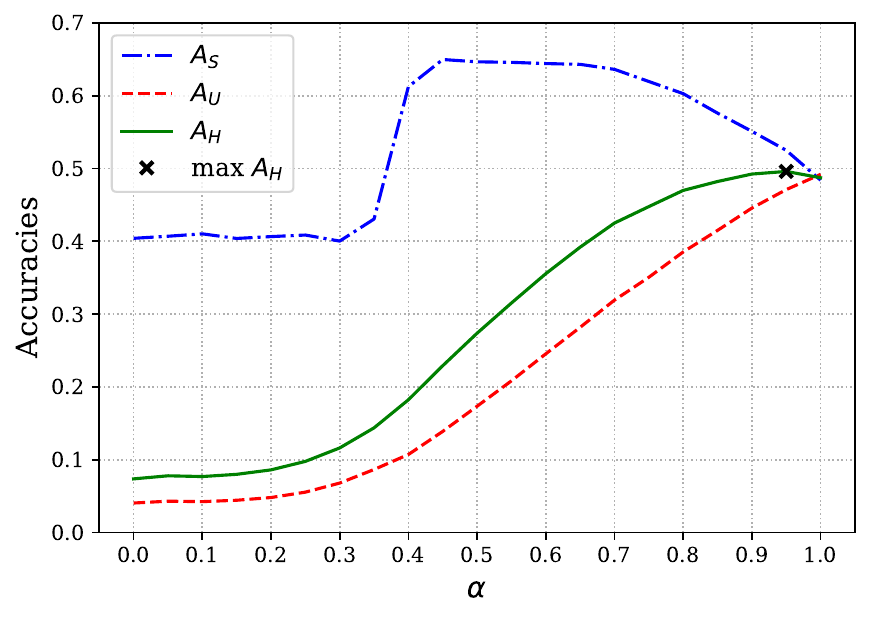}
            \caption[]%
            {{\small CUB}}    
            \label{}
        \end{subfigure}
        \begin{subfigure}[b]{0.45\textwidth}   
            \centering 
            \includegraphics[width=\textwidth]{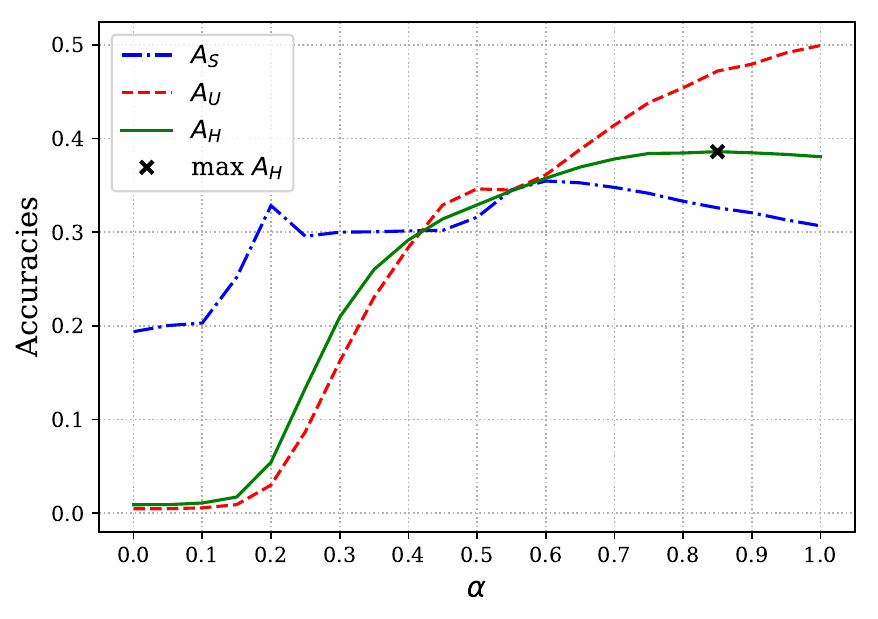}
            \caption[]%
            {{\small SUN}}    
            \label{}
        \end{subfigure}
        \caption[]
        {Plots of seen ($A_S$), unseen ($A_U$) and harmonic average ($A_H$) accuracies versus $\alpha$.  The maximum obtained harmonic accuracy is also marked by ($\times $).} 
        \label{fig:acc_vs_alpha}
        \vspace{-0.2in}
    \end{figure}

Moreover, SDM alleviates overconfidence to seen classes and introduces information about unseen classes during training phase. 
Figure \ref{fig:acc_vs_alpha} show the impact of $\alpha$ on seen ($A_S$), unseen ($A_U$) and harmonic average ($A_H$) accuracies. The plots represent that conventional cross-entropy loss ($\alpha=0$) results in almost zero unseen and harmonic average accuracies. This underscores the importance of similarity distribution regularization. As shown in Figure \ref{fig:acc_vs_alpha} for all datasets, not only unseen accuracy but also seen accuracy benefits from SDM, as the maximum of seen accuracy occurs at some $\alpha$ greater than zero which confirms the significance of probability distribution created by similarity values for cross-entropy loss in GZSL setting. 
%%%%%

It should be noted that both AwA1 and aPY datasets (Figures \ref{fig:acc_vs_q}.a, \ref{fig:acc_vs_q}.b, \ref{fig:acc_vs_alpha}.a ,\ref{fig:acc_vs_alpha}.b)  are coarse-grained class datasets. In contrast, CUB and SUN datasets are fine-grained with hundreds of classes and highly unbalanced seen-unseen split, and hence their accuracies have different behavior concerning $q$ and $\alpha$, as shown in Figures (\ref{fig:acc_vs_q}.c \ref{fig:acc_vs_q}.d \ref{fig:acc_vs_alpha}.c, \ref{fig:acc_vs_alpha}.d). However, harmonic average curve still has the same behavior and possesses a maximum value at an optimal $q$ and $\alpha$.

\begin{figure}
        \centering
        \begin{subfigure}[b]{0.45\textwidth}
            \centering
            \includegraphics[width=\textwidth]{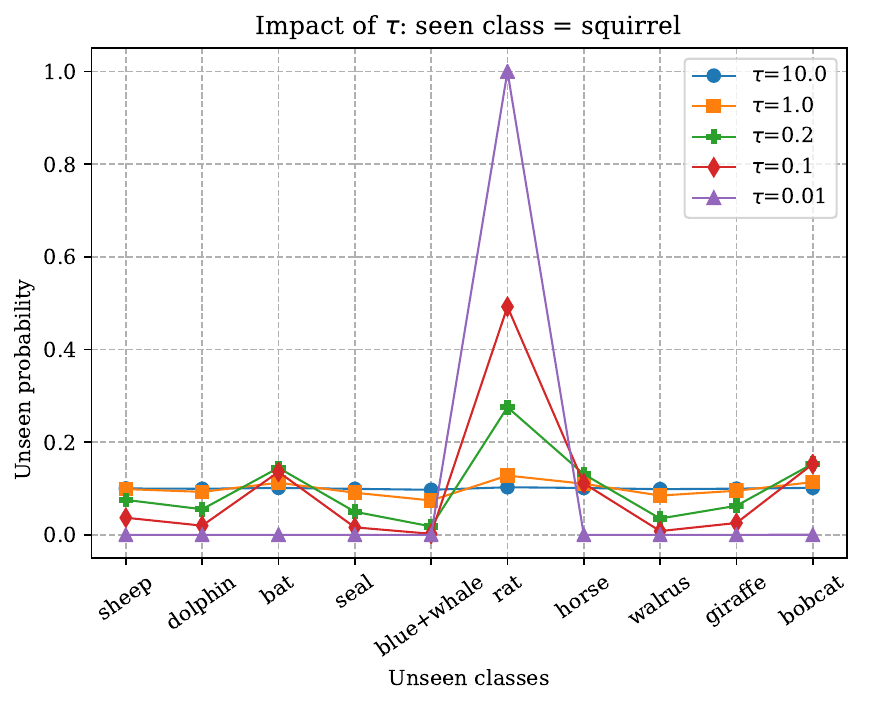}
            \caption[]%
            {{\small }}    
            \label{}
        \end{subfigure}
        \hfill
        \begin{subfigure}[b]{0.47\textwidth}  
            \centering 
            \includegraphics[width=\textwidth]{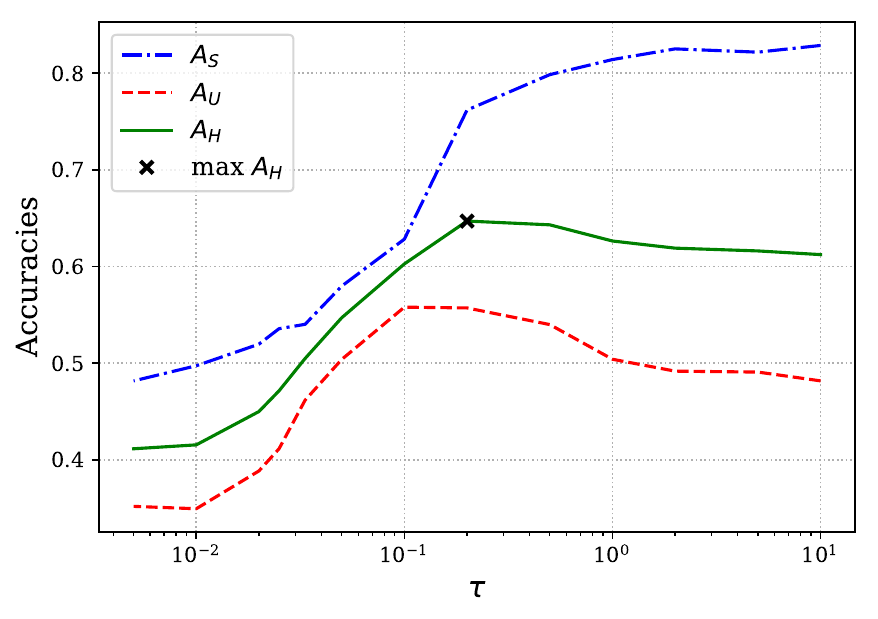}
            \caption[]%
            {{\small }}    
            \label{}
        \end{subfigure}
        \caption[]
        {The impact of temperature parameter $\tau$ for AwA1 dataset. \emph{left:} unseen probabilities (before multiplying $q$) produced by temperature Softmax Equation (\ref{eq:softl2}) for various $\tau$. \emph{right:} accuracies versus $\tau$ for proposed SDM classifier.}
        \label{fig:temp_effect}
        \vspace{-0.2in}
\end{figure}

\subsection{Intuition} 
We illustrate the intuition with AwA1 dataset ~\cite{lampert2009learning}. %and its proposed seen-unseen split \cite{goodbadugly}.%
Consider a seen class \emph{squirrel}. We compute closest unseen classes to the class \emph{squirrel} in terms of attributes. We naturally find that the closest class is \emph{rat} and the second closest is \emph{bat}, while other classes such as \emph{horse}, \emph{dolphin}, \emph{sheep}, etc. are not close (Figure \ref{fig:temp_effect}.\textit{left}). This is not surprising as \emph{squirrel} and \emph{rat} share several attribute. It is naturally desirable to have a classifier that gives \emph{rat} higher probability than other classes. If we force this softly, we can ensure that classifier is not blind towards unseen classes due to lack of any training example.  

From a learning perspective, without any regularization, we cannot hope classifier to classify unseen classes accurately. This problem was identified in ~\cite{liu2018generalized}, where they proposed entropy-based regularization in the form of Deep Calibration Network (DCN). DCN uses cross-entropy loss for seen classes, and regularizes the model with entropy loss on unseen classes to train the network. Authors in DCN postulate that minimizing the uncertainty (entropy) of predicted unseen distribution of training samples, enables the network to become aware of unseen visual features. While minimizing uncertainty is a good choice of regularization, it does not eliminate the possibility of being confident about the wrong unseen class.
% Clearly, in our example above, the uncertainty can be minimized even when the classifier gives high confidence to an unseen class \emph{dolphin} on an image of seen class \emph{squirrel}.
Clearly in DCN's approach, for the above \emph{squirrel} example, the uncertainty can be minimized even when the classifier gives high confidence to a wrong unseen class \emph{dolphin} on an image of seen class \emph{squirrel}.
%%%%remove????
%Furthermore, in many cases if several unseen classes are close to the correct class, we may not actually want low uncertainty.
Utilizing similarity distribution matching implicitly regularizes the model in a supervised fashion. The similarity values naturally has information of how much certainty we want for specific unseen class. We believe that this supervised regularization is the critical difference why our model outperforms DCN with a significant margin.

\subsection{Illustration of Similarity Distribution}
\label{sec:softlab} 
Figure \ref{fig:temp_effect} shows the effect of $\tau$ and the assigned unseen distribution on seen, unseen and harmonic accuracies for AwA1 dataset. Small $\tau$ enforces $q$ to be concentrated on nearest unseen class, while large $\tau$ spread $q$ over all the unseen classes and basically does not introduce helpful unseen class information to the classifier. The optimal value for $\tau$ is 0.2 for AwA1 dataset as depicted in Figure \ref{fig:temp_effect}.\textit{right}.  
The impact of $\tau$ on the assigned distribution for unseen classes is shown in Figure \ref{fig:temp_effect}.\textit{left} when seen class is \textit{squirrel} in AwA1 dataset. Unseen distribution with $\tau=0.2$, well represents the similarities between seen class (\textit{squirrel}) and similar unseen classes (\textit{rat}, \textit{bat}, \textit{bobcat}) and basically verifies the result of Figure \ref{fig:temp_effect}.\textit{right} where $\tau=0.2$ is the optimal temperature. While in the extreme cases, when $\tau=0.01$, distribution on unseen classes in mostly focused on the nearest unseen class, \emph{rat}, and consequently the other unseen classes' similarities are ignored. Also $\tau=10$ flattens the unseen distribution which results in high uncertainty and does not contribute helpful unseen class information to the learning.

\section{Conclusion}
We proposed a discriminative GZSL classifier with visual-to-semantic mapping and cross-entropy loss. During training, while SDM is trained on a seen class, it simultaneously learns similar unseen classes through probability distribution based on semantic similarity. We construct similarity distribution on unseen classes which allows us to learn both seen and unseen signatures simultaneously via a simple architecture. Our proposed similarity distribution matching strategy along with cross-entropy loss leads to a novel regularization via generalized similarity-based weighted cross-entropy loss that can successfully tackle GZSL problem. SDM offers a trade-off between seen and unseen accuracies and provides the capability to adjust these accuracies based on the particular application. We achieve competitive performance with state-of-the-art methods in GZSL setting, on all five ZSL benchmark datasets while keeping the model simple, efficient and easy to train.

\medskip

\small

% \bibliographystyle{unsrt}
% \bibliography{refs}

\end{document}